%% file: paper.tex
\documentclass[conference]{IEEEtran}
\usepackage{paper}
\def\BibTeX{{\rm B\kern-.05em{\sc i\kern-.025em b}\kern-.08em
    T\kern-.1667em\lower.7ex\hbox{E}\kern-.125emX}}
    
\makeatletter
\def\ps@IEEEtitlepagestyle{%
  \def\@oddfoot{\mycopyrightnotice}%
}
\def\mycopyrightnotice{%
\fbox{\parbox{\dimexpr\textwidth-2\fboxsep-2\fboxrule\relax}{
\begin{minipage}{\textwidth-2\fboxsep-2\fboxrule}
  \footnotesize
  \textcopyright 2022 IEEE. Personal use of this material is permitted. Permission from IEEE must be obtained for all other uses, in any current or future media, including reprinting/republishing this material for advertising or promotional purposes, creating new collective works, for resale or redistribution to servers or lists, or reuse of any copyrighted component of this work in other works.
  \end{minipage}
}}
}
\makeatother

\begin{document}

\title{Towards automating Numerical Consistency Checks in Financial Reports}

\author{\IEEEauthorblockN{Lars  Hillebrand\IEEEauthorrefmark{1}\IEEEauthorrefmark{2}\IEEEauthorrefmark{3}, Tobias Deußer\IEEEauthorrefmark{2}\IEEEauthorrefmark{3}, Tim Dilmaghani\IEEEauthorrefmark{4}, Bernd Kliem\IEEEauthorrefmark{4}, \\ Rüdiger Loitz\IEEEauthorrefmark{4}, 
Christian Bauckhage\IEEEauthorrefmark{2}\IEEEauthorrefmark{3}, Rafet Sifa\IEEEauthorrefmark{2}}
\IEEEauthorblockA{\IEEEauthorrefmark{2}\textit{Fraunhofer IAIS}, Bonn, Germany \\
\IEEEauthorrefmark{3}\textit{University of Bonn}, Bonn, Germany\\
\IEEEauthorrefmark{4}\textit{PricewaterhouseCoopers GmbH}, Düsseldorf, Germany
}}


\maketitle

\begin{abstract}
We introduce KPI-Check, a novel system that automatically identifies and cross-checks semantically equivalent key performance indicators (KPIs), e.g. ``revenue'' or ``total costs'', in real-world German financial reports. It combines a financial named entity and relation extraction module with a BERT-based filtering and text pair classification component to extract KPIs from unstructured sentences before linking them to synonymous occurrences in the balance sheet and profit \& loss statement. The tool achieves a high matching performance of $73.00$\% micro F$_1$ on a hold out test set and is currently being deployed for a globally operating major auditing firm to assist the auditing procedure of financial statements.
\end{abstract}


\section{Introduction}
Corporate disclosure documents like annual financial statements, management reports or initial public offering (IPO) prospectuses play a vital role in informing the public about a company's economic state of affairs. They contain large amounts of numerical facts that convey detailed information about profitability, financial strength and operational efficiency. These key performance indicators (KPIs) greatly affect investment decisions of outside investors and in return impact the company's future development. Hence, their authenticity, factual correctness and consistency within the report is of immense importance, which is reflected in strict reporting standards, e.g. IFRS (international financial reporting standards), whose compliance is regularly validated by external auditors. 

Companies themselves and external auditors spend a considerable amount of time to manually cross-check these numerical facts and financial indicators, which occur in tables and are further explained and referred to in various sections of text across the entire document. Due to their large volume and a tedious report generation process, which often involves multiple authors, frequent updates and copy-pasting, they are prone to numeric inconsistencies. These errors frequently persist even after official publishing, which negatively affects the reader's impression of the firm and thus, harms its reputation and integrity. Several studies have quantified the negative effect of accounting errors on investment decisions and the resulting economic consequences. For example, \cite{lawrence2013individual} and \cite{nwaobia2013financial} show that individual investors prefer to invest in companies with accessible and transparent disclosures. \cite{choudhary2021immaterial} and \cite{fang2017imperfect} go a step further and find that accounting errors are negatively associated with share returns and cause market participants to react less to earnings surprises. Therefore, reducing the amount of numerical errors in disclosure documents, while at the same time speeding up the tedious cross-checking process, is in the best interest of companies as well as auditing firms.

\begin{figure*}[t]
  \centering
  \includegraphics[width=0.75\textwidth]{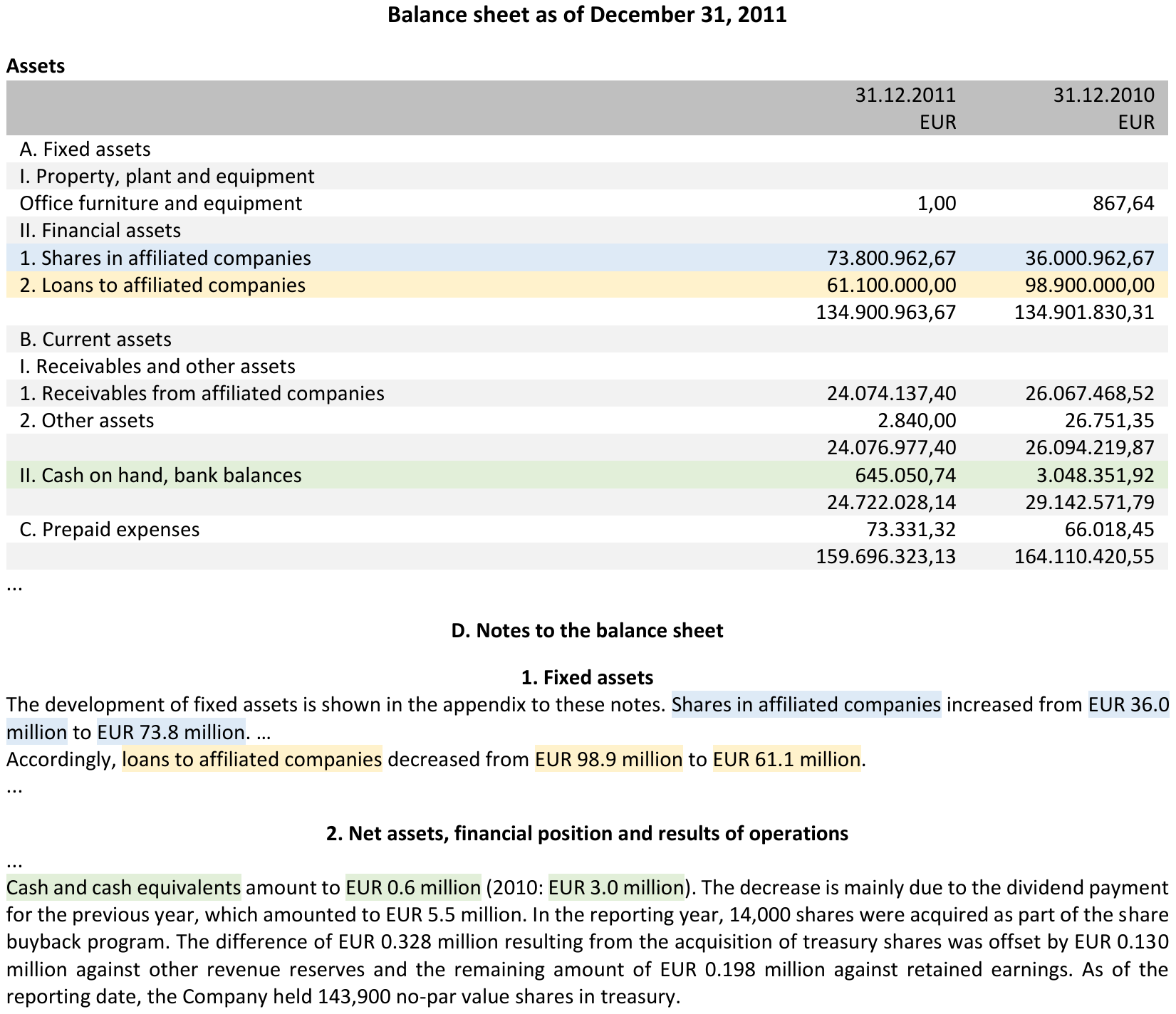}
  \caption{A screenshot of selected parts of a German financial statement (translated to English via DeepL, \url{https://www.deepl.com/}) showcasing the successful linking of semantically equivalent key performance indicators (KPIs) that occur in the balance sheet and at different places across the document text. KPIs colored equally refer to the same fact and thus, have to be numerically consistent. 
}
  \label{fig:report_example}
\end{figure*}

To tackle these objectives, we introduce KPI-Check, a sophisticated system that automatically identifies and cross-checks semantically equivalent KPIs in real-world German financial documents. 
Figure \ref{fig:report_example} shows an excerpt of such a document, in which textual KPIs and their numeric values are successfully identified across the document and matched with their balance sheet counterparts (equal color). Subsequently, the related pairs can be validated for numerical consistency by simply comparing their monetary values taking the scale (e.g. million) and unit (e.g. \euro{}) into account.

The balance sheet and profit \& loss statement arguably represent the most important sources of information within a financial report. Together, they can be used to assess the year-to-year consistency, performance and organizational direction of a company. That is why KPI-Check focuses on matching retrieved textual KPIs to these table types.

To achieve the non-trivial linking of synonymous KPIs, our tool consists of three dependent building blocks. 

First, we leverage KPI-BERT \cite{hillebrand2022kpi}, a novel named entity recognition and relation extraction model tailored to the financial domain, which jointly retrieves KPIs from sentences and relates them to their numeric values. Given the following example sentence,
\begin{tcolorbox}[breakable,notitle,boxrule=0pt,
boxsep=0pt,left=0.6em,right=0.6em,top=0.5em,bottom=0.5em,
colback=gray!10,
colframe=gray!10]
\small
``In 2021 the $\underset{\text{{\color{ForestGreen}{kpi}}}}{\text{{\color{ForestGreen}{revenues}}}}$ increased from \$$\underset{\text{{\color{ForestGreen}{py}}}}{\text{{\color{ForestGreen}{76}}}}$ million to \$$\underset{\text{{\color{ForestGreen}{cy}}}}{\text{{\color{ForestGreen}{112}}}}$ million while the $\underset{\text{{\color{ForestGreen}{kpi}}}}{\text{{\color{ForestGreen}{total costs}}}}$ decreased to \$$\underset{\text{{\color{ForestGreen}{cy}}}}{\text{{\color{ForestGreen}{47}}}}$ million  (prior year: \$$\underset{\text{{\color{ForestGreen}{py}}}}{\text{{\color{ForestGreen}{66}}}}$ million).''

\vspace{0.2cm}
{\small
$\underset{\text{{\color{ForestGreen}{kpi}}}}{\text{{\color{ForestGreen}{revenue}}}} - \underset{\text{{\color{ForestGreen}{cy}}}}{\text{{\color{ForestGreen}{112}}}}$, $\underset{\text{{\color{ForestGreen}{kpi}}}}{\text{{\color{ForestGreen}{revenue}}}} - \underset{\text{{\color{ForestGreen}{py}}}}{\text{{\color{ForestGreen}{76}}}}$, $\underset{\text{{\color{ForestGreen}{kpi}}}}{\text{{\color{ForestGreen}{total costs}}}} - \underset{\text{{\color{ForestGreen}{cy}}}}{\text{{\color{ForestGreen}{47}}}}$,
$\underset{\text{{\color{ForestGreen}{kpi}}}}{\text{{\color{ForestGreen}{total costs}}}} - \underset{\text{{\color{ForestGreen}{py}}}}{\text{{\color{ForestGreen}{66}}}}$}
\end{tcolorbox}
it automatically recognizes and classifies the highlighted financial indicators and links their numeric relations, 
where \textit{kpi}, \textit{cy} (current year value) and \textit{py} (prior year value) are part of previously defined entity classes. 

Second, we integrate a joint sentence- and table encoding module which utilizes ``Bidirectional
Encoder Representations from Transformers'' (BERT) \cite{devlin2018bert} to find relevant sentence/table pairs within a financial report. One of the key challenges of linking semantically equivalent KPIs is the extreme data imbalance, since the large majority of KPI-pair combinations are unrelated. Hence, this filtering module substantially simplifies the matching task, which positively impacts the system's performance.

Lastly, we introduce a binary prediction module using a contrastive autoencoder (CAE) to classify the remaining KPI-pairs employing weighted sampling techniques to expose related pairs more frequently during training.

The complete system achieves a final test set micro F$_1$-score of $73.00$\%, which shows the model's capability to learn semantic similarities despite the task's difficulty and the aforementioned imbalance challenge. 

KPI-Check is currently being deployed for a major auditing firm as a separate component of an AI-based auditing tool for financial statements. It adds a convenient method to automatically retrieve and highlight identical KPIs and to detect numerical inconsistencies in financial documents. First user tests have already revealed significant efficiency gains and the continuous use in production will further improve the system's performance due to the integration of human feedback, e.g. in the form of error corrections.


\begin{figure*}[t]
  \centering
  \includegraphics[width=\linewidth]{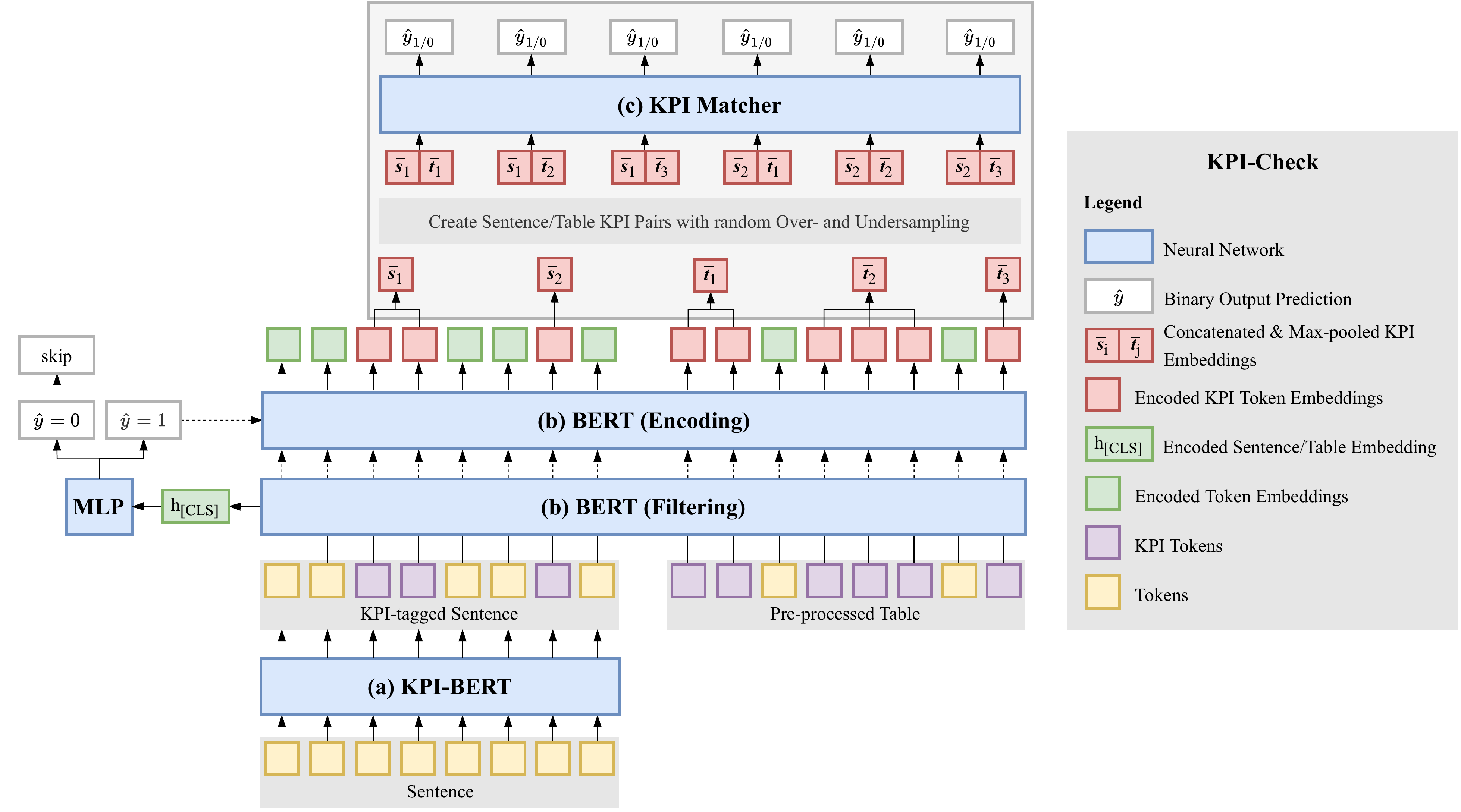}
  \caption{Schematic visualization of our system, KPI-Check, to automatically identify and link semantically equivalent key performance indicators (KPIs) in sentences and tables within financial documents. First, (a) sentences are passed through KPI-BERT, a joint named entity and relation extraction model, to retrieve KPIs and their numeric values. Second, (b) a BERT-based filtering model using cross-attention with a multi-layer perceptron (MLP) classification head classifies a KPI-tagged sentence and pre-processed table pair to either match (contain equivalent KPIs) or not. If a match is predicted, a separate BERT-based encoding module utilizing max-pooling creates encoded sentence- and table KPI embeddings, $\mat{\bar{s}}$ and $\mat{\bar{t}}$ . Finally, (c) a contrastive autoencoder (CAE) classifies each sentence/table KPI-pair to be synonymous or not.}
  \label{fig:kpi-check}
\end{figure*}

\section{Related Work}
\label{sec:related}


In today's digitized world with an ever-growing amount of freely available information claim-checking becomes more and more important to guarantee factual correctness and consistency. Over the past years, it has sparked much research interest across multiple disciplines. For example, \cite{nuijten2016statcheck} introduce StatCheck, a rule-based tool that detects inconsistencies during significance testing in academic psychology papers. \cite{hassan2017claimbuster} develop a fact-checking platform called ClaimBuster that utilizes natural language processing and supervised learning to identify important factual claims in political discourses. Additionally, \cite{kobayashi2017automated} and \cite{nadeem2019fakta} present end-to-end fact checking systems that predict the factuality of historically given claims.

The majority of these systems rely on pre-trained natural language models (\cite{devlin2018bert}, 
\cite{Liu2019RoBERTaAR}, 
\cite{Radford2018ImprovingLU}) and utilize named entity recognition (\cite{parolin2021come},
\cite{wang2021ImprovingNE}, 
\cite{wang2020automated})
\cite{Yamada2020LUKE}, 
\cite{Yu2020NamedER}) and relation extraction (\cite{crone2020deeper}, 
\cite{eberts2019span}, 
\cite{Shen2021ATM}, 
\cite{Wang2020TwoAB},
\cite{Ye2021PackTE}, 
\cite{Zhong2021AFE}), of which KPI-Check is no exception since it makes use of KPI-BERT \cite{hillebrand2022kpi}, a  joint named entity and relation extraction model tailored to the financial domain. 

Turning to our concrete task of analyzing and cross-checking financial documents, \cite{xu2021jura} suggest Jura, a machine-learning based compliance tool to improve the efficiency of reviewing annual financial reports submitted to the Hong Kong Exchanges and Clearing (HKEX). Similarily, \cite{sifa2019towards} and \cite{ramamurthy2021alibert} propose and improve ALI, a recommender-based tool that greatly simplifies and to a large extend automates the auditing of financial statements. 

However, the previously named systems lack our focus on numerical consistency of key performance indicators (KPIs). The closest research in this regard are probably the studies by \cite{Cao2018TowardsAN} and \cite{li2020cracking}. 
The former extracts formulas from verbal descriptions of numerical claims while leaving the actual claim linking task for future work. Also, they extract financial indicators using a whitelist, which might not be general enough depending on the variety of KPIs. The latter study analyzes how well different KPIs in tables can be cross-checked in Chinese IPO prospectuses and auditing reports. The authors achieve great results for identifying semantically equivalent table cells, but we consider it problematic that they leak numerical information in the matching process, which they claim is purely based on semantics.

Since identifying semantically equivalent KPI pairs in financial reports suffers from an extremely high data imbalance KPI-Check's task qualifies as an outlier/anomaly detection problem. In the past many different architectures and training schemes have been proposed to tackle such problems. However, autoencoders (\cite{adkisson2021autoencoder}, 
\cite{chen2020autoencoder}, 
\cite{li2021vaga}, 
\cite{lubbering2021supervised}, 
\cite{lubbering2020imbalanced}) stand out in popularity, both in the supervised and unsupervised setting. KPI-Check, being no exception, leverages a contrastive autoencoder (CAE) that uses contrastive learning on the reconstruction loss to separate equivalent from unrelated KPI pairs. 

\section{Methodology} \label{sec:method}
In this section, we briefly formulate the problem and motivate our modeling approach before turning to the in-depth analysis of our proposed architecture which is visualized in Figure \ref{fig:kpi-check}.

\subsection{Problem Formulation and Modeling Approach}

Given a corporate financial report containing a list of tabular key performance indicators (KPIs), $\mathcal{T}$, depicted in the balance sheet and profit \& loss statement, and a list of sentence KPIs, $\mathcal{S}$, occurring across the entire document, we identify all semantically equivalent pairs from the combined Cartesian product $\mathcal{T} \times \mathcal{S}$. Once we succeed in this task, we can automatically cross-check their monetary values and thus, verify numerical consistency.

Effectively, the above described objective can be divided into three sub-problems. 

First, we extract the originally unknown performance indicators, $\mathcal{T}$ and $\mathcal{S}$, along with their numerical quantities from the document. In the case of $\mathcal{T}$ this can be done rule-based, due to the known structure of the balance sheet and profit \& loss statement. On the contrary, retrieving $\mathcal{S}$ is more difficult which is why we utilize KPI-BERT \cite{hillebrand2022kpi}, a dedicated named entity and relation extraction model that is trained to detect and link KPIs within unstructured sentences (see Figure \ref{fig:kpi-check}a). KPI-Check uses the extracted and linked KPIs subsequently in the following two tasks.

Second, we create vector representations (embeddings) for all extracted KPIs in $\mathcal{T}$ and $\mathcal{S}$ that capture their semantics and contextual information (see Figure \ref{fig:kpi-check}b). It is important to note that we explicitly exclude the KPI's numeric quantity from the embedding process so that our classification module is forced to only learn semantic and contextual similarities instead of focusing on numerical equivalence. Enabling the latter would open up the possibility to link KPIs solely based on their monetary values, which contradicts our idea of matching identical KPIs to find numerical inconsistencies. 

Third and finally, we classify each KPI-pair $(t, s) \in \mathcal{T} \times \mathcal{S}$ to either \textit{match}, $+$, or \textit{not match}, $-$ (see Figure \ref{fig:kpi-check}c). Due to the huge discrepancy in the amount of synonymous (small) and unrelated (large) KPIs, a key challenge is the extremely high classification imbalance of $\frac{\lvert \mathcal{T} \times \mathcal{S} \rvert^-}{\lvert \mathcal{T} \times \mathcal{S} \rvert^+}\approx 500:1$ which effectively qualifies this task as outlier detection problem. To reduce this issue, we introduce a separate filtering module which is trained to remove irrelevant sentence/table pairs not containing any matching KPIs before performing the actual KPI linking step. In addition, we incorporate class-weighted sampling in the training process to expose the minority class of matching KPI-pairs more frequently. Lastly, we perform the final KPI linking step by introducing a contrastive autoencoder (CAE) that utilizes contrastive learning to robustly differentiate between inliers (unrelated KPIs) and outliers (synonymous KPIs).

The following sections describe our solutions to these sub-tasks in more detail.

\subsection{Entity Extraction and Relation Linking (KPI-BERT)} \label{sec:kpi-bert}

To retrieve and connect all textual KPIs, $\mathcal{S}$, and their numeric quantities from unstructured sentences in a financial report, we leverage a named entity and relation extraction model, called KPI-BERT \cite{hillebrand2022kpi}. The model consists of three stacked components that are trained jointly in an end-to-end fashion via gradient descent. For completeness, we briefly summarize KPI-BERT's sub-modules while referring to \cite{hillebrand2022kpi} for all details.

\subsubsection{Sentence Encoder} \label{sec:sent-encoder}

A pre-trained BERT \cite{devlin2018bert} model encodes a WordPiece \cite{schuster2012japanese} tokenized sentence to obtain a contextualized sub-word token embedding sequence. Next, a bidirectional gated recurrent unit (Bi-GRU) pooling function is applied to create word representations $\mat{e}$. 

\subsubsection{Named Entity Recognition Decoder}

A GRU-based named entity recognition (NER) tagger sequentially retrieves KPIs and their numeric values leveraging conditionally masked label transitions and learnable entity type embeddings. Each encoded and pooled word embedding $\mat{e}_j$ is concatenated with the trainable embedding $\mat{w}_{j-1}^\text{label}$ of the previously predicted entity type, yielding the input representation of word $j$, $\mat{z}_j = \left[ \mat{e}_j; \mat{w}_{j-1}^\text{label}\right]$. 

Along with the previous hidden state $\mat{h}_{j-1}$, $\mat{z}_j$ is sequentially passed into a GRU, yielding $\mat{h}_j = \text{GRU}\left( \mat{z}_j, \mat{h}_{j-1}\right)$. To get an entity tag prediction for word $j$, $\mat{h}_j$ is linearly transformed, followed by masking out impossible entity types and applying softmax:
\begin{align}
    \mat{\hat{y}}_j = \text{softmax}\left( \text{mask}\left(\mat{W}_{\text{seq}} \mat{h}_j + \mat{b}_{\text{seq}}\right) \right).
\end{align}
Masking is applied conditional on the previously predicted tag $\argmax(\mat{\hat{y}}_{j-1})$, i.e. following an \textit{O} (outside) tag no \textit{I} (inside) or \textit{E} (end) can occur. For more details on the conditional masking logic we refer to Section III-B in \cite{hillebrand2022kpi}.

\subsubsection{Relation Extraction Decoder}

The tagged words and their embeddings $\mat{e}$ are converted to the entity-level by applying the same BI-GRU pooling function as above and concatenated to a trainable size embedding $\mat{w}_{k}^\text{width}$. Hence, entity $s$ of size $k$ is represented as
\begin{equation}
\mat{e}(s) := \left[\text{Bi-GRU}_\text{pool}(\mat{e}_j, \mat{e}_{j+1}, \dots, \mat{e}_{j+k-1});\mat{w}_{k}^\text{width}\right].
    \label{eq:span_representation}
\end{equation}

To relate KPIs to their numeric values, candidate pairs $(s_i, s_j)$ are sampled from the pool of \textit{allowed} entity combinations in the sentence. Given two entities, $s_1$ and $s_2$, their respective representations are concatenated with a localized context embedding $\mat{c}_{\text{loc}}$, which contains the pooled word embeddings located between $s_1$ and $s_2$.
Hence, $\mat{x}_r(s_1, s_2) := \left[\mat{e}(s_1);\mat{c}_{\text{loc}}(s_1, s_2); \mat{e}(s_2) \right]$
is defined as input for the relation classifier, which can be formalized as
\begin{equation}
    \hat{y}_r = \text{sigmoid} \left( \mat{w}_{\text{rel}}^T \mat{x}_r(s_1, s_2)  + b_{\text{rel}} \right),
    \label{eq:relation_classifier}
\end{equation}
where $\mat{w}_{\text{rel}}$ and $b_{\text{rel}}$ are weight and bias terms, respectively.
If $\hat{y}_r \in [0,1]$ exceeds pre-defined confidence threshold, we consider entity $s_1$ and entity $s_2$ to match.

\subsection{Entity Encoding}

After KPI-BERT \cite{hillebrand2022kpi} successfully extracts all textual KPIs in $\mathcal{S}$ and links their numeric quantities within sentences, we turn to the second sub-task of encoding sentence- and table KPIs into vector space while preserving their semantics and context. 

\subsubsection{Pre-Processing} \label{sec:pre-processing}
For the balance sheet and profit \& loss statement we only regard their first column, which comprises the entirety of tabular KPIs, $\mathcal{T}$. The current and prior year numeric values, present in the remaining table columns, are deliberately discarded in the embedding process. Including them would potentially enable the downstream classification model to base its predictions mainly on the numerical equivalence of two KPIs. This would be problematic since semantically identical KPIs with inconsistent numerical values, e.g. caused by human error, might not be detected.
\begin{figure}[t]
 \centering
 \begin{subfigure}[b]{\linewidth}
     \footnotesize
     \centering
     \input{tables/pandl-example}
     \caption{Excerpt of a profit \& loss statement.}
     \label{tab:pandl-example}
 \end{subfigure}
 \par\bigskip
 \begin{subfigure}[b]{\linewidth}
     \centering
     \begin{tcolorbox}[breakable,notitle,boxrule=0pt,
        boxsep=0pt,left=0.6em,right=0.6em,top=0.5em,bottom=0.5em,
        colback=gray!10,
        colframe=gray!10]
        \footnotesize
        Gross profit $<$row$>$  $<$nan$>$  $<$row$>$  Personnel expenses $<$row$>$  Wages and salaries $<$row$>$  Social security contributions $<$row$>$  \dots
     \end{tcolorbox}
     \caption{Flattened and pre-processed profit \& loss statement.}
     \label{fig:pandl_processed}
 \end{subfigure}
 \caption{Example of a profit \& loss statement (a) and its pre-processed and flattened key performance indicator (KPI) sequence (b).}
 \label{fig:tab-example}
\end{figure}

We flatten and pre-preprocess each table, which is exemplary depicted in Figure \ref{fig:tab-example}. First, we use rule-based heuristics and regular expressions to remove àny hierarchical prefix, i.e. converting ``1. Gross profit'' to ``Gross profit''. Second, we include two special tokens, $<$row$>$ and $<$nan$>$ to separate new rows and tag empty KPIs, respectively.

To prepare the previously extracted sentence KPIs for encoding, we process each sentence by enclosing all tagged KPIs with HTML-like special tokens, i.e.
 \begin{tcolorbox}[breakable,notitle,boxrule=0pt,
        boxsep=0pt,left=0.6em,right=0.6em,top=0.5em,bottom=0.5em,
        colback=gray!10,
        colframe=gray!10]
        \small
        In 2022, the $<$kpi$>$ revenue $<$/kpi$>$ increased to \$1.5 million.
\end{tcolorbox}

\subsubsection{Filtering}
As visualized in Figure \ref{fig:kpi-check}b, a filtering module decides prior to linking KPIs whether a candidate sentence/table pair contains synonymous KPIs. Concretely, we employ a pre-trained and in the process fine-tuned BERT model which jointly encodes the processed sentence and table using cross-attention to learn a combined sentence/table representation
\begin{align*}
    \mat{h}_\text{[CLS]} = \text{BERT}_\text{filter}\left( \text{[CLS] sentence [SEP] table [SEP]}\right),
\end{align*}
where [CLS] and [SEP] denote BERT-specific special tokens used for input classification and separation. Subsequently, we classify the pair's relevance by passing $\mat{h}_\text{[CLS]}$ to a simple multi-layer perceptron (MLP) consisting of a fully-connected layer followed by dropout and a sigmoidal activation function:
\begin{align}
    \hat{y}_\text{filter} = \text{MLP}_\text{filter}\left(\mat{h}_\text{[CLS]}\right). \label{eq:mlp}
\end{align}
If $\hat{y}_\text{filter} \in [0,1]$ is below a pre-defined confidence threshold $\alpha_1$, we discard the sentence/table pair to reduce the aforementioned data imbalance problem and thus, increase the final KPI matching performance.

\subsubsection{Encoding}
Similar to Section \ref{sec:sent-encoder} we use a dedicated pre-trained BERT model to separately encode pre-processed sentences and tables that succeeded the previous filtering process. Specifically, given a KPI-tagged sentence of $n$ tokens and a flattened table of $m$ tokens, we obtain context-aware sub-word embeddings:
\begin{align}
    \mat{s}_1, \mat{s}_{2}, \dots, \mat{s}_n &= \text{BERT}_\text{encode}\left(\text{[CLS] sentence [SEP]}\right), \\
    \mat{t}_1, \mat{t}_{2}, \dots, \mat{t}_m &= \text{BERT}_\text{encode}\left(\text{[CLS] table [SEP]}\right).
\end{align}
Next, we utilize the known positions of our KPIs within the embedding sequences and apply max-pooling to create KPI embeddings for both, tables and sentences. Similar to Equation \ref{eq:span_representation}, an arbitrary KPI in a sentence containing $k$ sub-words is represented as
\begin{align}
    \mat{\bar{s}} = \left[\text{max-pool}(\mat{s}_i, \mat{s}_{i+1}, \dots, \mat{s}_{i+k-1});\mat{w}^\text{width}_k\right],
    \label{eq:pooling}
\end{align}
where $\mat{w}^\text{width}_k$ again denotes a unique size embedding. $\mat{\bar{t}}$ follows the same approach for tables.

\subsection{Entity Pair Classification}

We cast the entity pair classification task as an outlier detection problem to further tackle the imbalance of relatively few synonymous KPI pairs compared to many unrelated KPI pairs. Specifically, we implement the KPI matching network depicted in Figure \ref{fig:kpi-check}c as a contrastive autoencoder (CAE) leveraging contrastive learning on the reconstruction loss to distinguish synonymous from unrelated pairs. The autoencoder is defined as
\begin{align}
    \varphi(\mat{x}) \coloneqq \text{dec}(\text{enc}(\mat{x})), \qquad \mat{x} = \left[\mat{\bar{s}};\mat{\bar{t}}\right],
\end{align}
where $\mat{x}$ represents the concatenation of the sentence- and table KPI embeddings $\mat{\bar{s}}$ and $\mat{\bar{t}}$ that are randomly sampled from the pool of pairs $\mathcal{T} \times \mathcal{S}$. The encoder (enc) and decoder (dec) networks are MLPs with two fully-connected layers each enclosed by relu activation functions and dropout. Following the standard design of autoencoders the hidden dimension imposes an information bottleneck ($\text{dim}_{\text{hidden}} \ll \text{dim}_{\text{input}} = \text{dim}_{\text{output}}$), which enforces meaningful representation learning for correct input reconstruction.

Given the reconstructed input, $\mat{\hat{x}} = \varphi(\mat{x})$, the original input $\mat{x}$ and the ground truth label $y \in \{0, 1\}$ we train the CAE to minimize the reconstruction loss for unrelated KPI-pairs, $y=0$, while maximizing it for semantically equivalent pairs, $y=1$. Concretely, we define the combined contrastive loss as
\begin{equation}
\begin{aligned}
    \mathcal{L}_{\text{contrastive}}(\mat{\hat{x}}, \mat{x}, y) = \: &(1 - y) \cdot \mathcal{L}_{\text{MSE}}(\mat{\hat{x}}, \mat{x}) \\ &+ y \cdot \text{max}(0,m - \mathcal{L}_{\text{MSE}}(\mat{\hat{x}}, \mat{x})), \label{contrastive}
\end{aligned}
\end{equation}
where $\mathcal{L}_{\text{MSE}}$ denotes the mean-squared error loss function and $m$ denotes the margin parameter enforcing that the model focuses particularly on difficult-to-reconstruct samples during optimization.

During inference we normalize the resulting mean-squared error loss with a sigmoid layer such that
\begin{align}
    \hat{y} = \text{sigmoid}(\mathcal{L}_{\text{MSE}}(\mat{\hat{x}}, \mat{x}))) \in [0,1].
\end{align}
If $\hat{y}$ is above the threshold $\alpha_2$, we consider the KPI-pair $(t,s)$ to be semantically equivalent.





\subsection{Training}

We decouple the training process of KPI-Check and train each sub-module independently due to the significantly different objectives in their sub-tasks and a more effective usage of tailored random over and undersampling (ROUS) per task.

First, our named entity and relation extraction model for sentences, KPI-BERT, trains on a smaller subset of 500 manually annotated financial reports\footnote{The reports are part of the training split for the other KPI-Check modules.}, to extract and link KPIs and their numeric quantities from each sentence. The model jointly optimizes the named entity (categorical cross entropy) and relation extraction (binary cross entropy) loss. Further details about the training and hyperparameter tuning process can be again taken from \cite{hillebrand2022kpi}. In this work we leverage the best performing version of KPI-BERT as reported in \cite{hillebrand2022kpi}.

Second, we train the components of the filtering module, $\text{BERT}_\text{filter}$ and $\text{MLP}_\text{filter}$, end-to-end while fine-tuning $\text{BERT}_\text{filter}$ explicitly on the task of identifying relevant sentence/table pairs. We optimize the binary cross entropy loss over related and unrelated pairs. Since most sentences and tables are unrelated and only a few share semantically equivalent KPIs, we employ weighted random over- and undersampling (ROUS) with replacement to show relevant pairs more often during training. Specifically, the originally uniform sampling probability of each sentence/table pair is altered to the normalized inverse frequency of the pair's class occurrence in the training set. For example, a training set of four pairs $(+,-,-,-)$ receives weighted sampling probabilities of $\left(\frac{1}{2}, \frac{1}{6}, \frac{1}{6}, \frac{1}{6}\right)$.

Third, we train the actual KPI-matching model, the CAE, which takes the encoded (BERT$_\text{encode}$) and pooled KPI representations as input, provided their sentence/table pair overcame the previous filtering process. We employ contrastive learning (see Equation \eqref{contrastive}) to learn a robust decision boundary which effectively distinguishes between unrelated and synonymous KPI-pairs. In addition, we leverage teacher forcing (training only) by also utilizing positively annotated KPI-pairs that were falsely filtered out. 

\section{Experiments} \label{sec:experiments}
In the following sections, we introduce our custom dataset of German financial reports, describe the automated annotation process of synonymous KPIs, discuss the overall training setup including hyperparameter optimization and evaluate results.

\subsection{Data}

Our dataset\footnote{We are currently unable to publish the dataset and the accompanying python code because both are developed and used in the context of an ongoing industrial project.} comprises $7,548$ real-world corporate annual reports sourced from the \textit{Bundesanzeiger}\footnote{\url{https://www.bundesanzeiger.de/}}, a platform hosted by the DuMont media group where German companies publish their legally required documents. A subset of $500$ reports (part of the training set) was utilized in \cite{hillebrand2022kpi} to train KPI-BERT, our joint named entity and relation extraction model that identifies and links KPIs in sentences. For the remainder of this paper we take a trained KPI-BERT model as a given and only consider it in the context of KPI-Check. For detailed information about annotations, model training and performance results we again refer to \cite{hillebrand2022kpi}.

We pre-process each report by first tokenizing on a sentence level and subsequently on a word level using the \texttt{syntok} python library. Second, we tag monetary numbers and extract their scale (e.g. million) and unit (e.g. \$) using regular expressions. Similarly, we utilize rule-based string matching heuristics and trigger words to identify the balance sheet and profit \& loss statement within each document. Third, we ignore all other tables and discard sentences not containing any numeric quantities because our only interest lies in linking KPI entities to check their numerical consistency. 

For model training, tuning and evaluation purposes we randomly divide the dataset into a fixed split of $6,032$ training, $764$ validation and $752$ testing documents. Table \ref{tab:data-statistics} presents detailed KPI- and sentence/table pair statistics for all splits. Each document contains on average around $8$ semantically equivalent and $2,900$ unrelated KPI-pairs leading to a per-document imbalance of around $500:1$. The overall imbalance across the entire training set still amounts to $367$ negative pairs for each positive one, which emphasizes the need for a filtering approach. The corresponding model, BERT$_\text{filter}$, requires sentence/table inputs, whose class ratio of negative ($-$) to positive ($+$) pairs is significantly lower equaling $12:1$.

\begin{table}[t]
     \centering
     \input{tables/data-statistics}
     \caption{Dataset statistics about key performance indicator (KPI) and sentence/table pairs, highlighting their respective class imbalances of positive ($+$) and negative ($-$) pairs.}
     \label{tab:data-statistics}
 \end{table}
 
\subsection{Automated Annotation Process} \label{sec:number-matching}
Manually cross-checking over $7,000$ financial documents to get ground truth annotations for semantically equivalent KPI-pairs would be extremely time-consuming and is practically infeasible in the scope of our project . Hence, we introduce an automated annotation logic which leverages informed number matching to efficiently annotate synonymous KPIs while keeping the amount of false positive and false negative annotations to a minimum.

As described in Section \ref{sec:kpi-bert} we identify and link each KPI within a sentence to its numeric quantities, i.e. current year (\textit{cy}) and prior year (\textit{py}) values. In addition, the value's unit (e.g. \euro{}) and scale (e.g. million) is accurately extracted employing regular expressions and rule based heuristics. The same is true for KPIs occurring in the balance sheet and profit \& loss statement. They can be automatically aligned with their quantities due to the known tabular structure.

Utilizing the accurate linking of each KPI to its numerical quantity, we apply informed number matching taking a number's unit, scale and potential rounding into account. For example, our matching heuristic recognizes that the tabular value of ``EUR $73,800,962.67$'' and the textual value of ``EUR $73.8$ million'' match (see Figure \ref{fig:report_example}).

Of course, relying on number matching for linking semantically equivalent KPIs is not perfect. In particular, two sources of noise might dilute the annotation quality. First, all equivalent KPIs whose numeric quantities do not match, i.e. caused by human error due to wrong rounding or copy-pasting, remain undetected (false negative annotations). Second, two numbers can match by chance although their KPIs are completely unrelated (false positive annotations). 

The first issue cannot be avoided but occurs presumably quite rarely. The second issue can be mitigated by applying the following rules.

If two candidate KPIs are respectively linked to current year and prior year values (\textit{cy} and \textit{py}), we numerically compare both quantities. In case of two matches, we confidently label the KPI-pair positively.
If one of the candidate KPIs is linked to only a single numeric quantity (either \textit{cy} or \textit{py}), we can only focus on comparing this quantity. In case of a numerical match we additionally confirm whether the matched number $x$ is ``common'' by validating if the decade logarithm of the absolute value of $x$ returns a natural number, i.e. $\text{log}_{10}(\lvert x\rvert) \in \NN$. If the previous expression evaluates to \texttt{False} (not common) we directly label the KPI-pair positively. If it evaluates to \texttt{True} we add an extra safety cushion by applying fuzzy string matching on the raw strings of both candidate KPIs. Concretely, we use the Levenshtein distance \cite{levenshtein1965binary} based weighted ratio function \texttt{WRatio()} of the \texttt{rapidfuzz} Python library to check whether the character similarity of both KPIs is below or above a strict threshold of $0.9$ (similarity is bound between $0$ and $1$). Only if the textual similarity is $\geq 0.9$ we label the candidate KPI-pair positively. In all other cases pairs are labeled negatively.

We extend this logic to the sentence/table level, by annotating a corresponding pair positively, if at least two KPIs within the pair match.

A qualitative analysis of randomly sampled annotations supports our hypothesis of a high labeling accuracy. Out of $100$ positively labeled samples, $97$ were actually correct. Also, a small amount of annotation noise is acceptable due the large amount of data and the proven robustness of deep neural networks \cite{rolnick2017deep}. In addition, employing the available numerical information for automated labeling is fully decoupled from the actual modeling process. As described in Section \ref{sec:pre-processing}, KPI-Check discards all numerical information and solely uses semantics and context to learn useful representations for synonymous KPI matching.

In Section \ref{sec:results} we see that KPI-Check generalizes well and is indeed able to predict correct KPI matches that were missed during the automated annotation process.

\subsection{Evaluation Metrics}

We quantitatively evaluate our system's performance by calculating precision, recall and F$_1$ scores. For a single document $d$ and given sets of predicted- and ground truth KPI-pairs, $\hat{\mathcal{Y}}$ and $\mathcal{Y}^*$, the three metrics are defined as
\begin{align*}
    \text{Precision} &= \frac{\lvert \hat{\mathcal{Y}} \cap \mathcal{Y}^* \rvert}{\lvert \hat{\mathcal{Y}} \rvert}, \quad
    \text{Recall} = \frac{\lvert \hat{\mathcal{Y}} \cap \mathcal{Y}^* \rvert}{\lvert  \mathcal{Y}^* \rvert}, \\
    \text{F}_1 &= 2 \cdot \frac{\text{Precision} \cdot \text{Recall}}{\text{Precision} +  \text{Recall}}.
\end{align*}

For $N$ documents ($d_1, d_2, \dots, d_N$) we calculate macro scores by averaging the document-level metrics and micro scores by aggregating $\hat{\mathcal{Y}}$ and $\mathcal{Y}^*$ across all documents. For example, macro and micro recall are respectively defined as
\begin{align*}
    \text{Recall}_\text{macro} = \frac{\sum_{i=1}^N \text{Recall}_i}{N}, \quad
    \text{Recall}_\text{micro} = \frac{\sum_{i=1}^N \lvert \hat{\mathcal{Y}}_i \cap \mathcal{Y}^*_i \rvert}{\sum_{i=1}^N \lvert \mathcal{Y}^*_i \rvert}, 
\end{align*}
where the subscript $i$ refers to the $i$-th document. Macro and micro metrics for precision and F$_1$ score are calculated analogously.

\subsection{Training Setup}

In this section, we shed light on the training setup and hyperparameter optimization of the individual components of KPI-Check (with the exception of KPI-BERT, which is thoroughly described in \cite{hillebrand2022kpi}).

We determine the best hyperparameter setup for each sub-module by conducting an extensive grid search analyzing various parameter combinations based on their validation set micro F$_1$-score. Table \ref{tab:grid_search} shows all tuned model parameters with their respective ranges of values. The best performing parameter setup on the validation set is highlighted in boldface.

Each sub-module employs the cased BERT$_{\text{BASE}}$ encoder,
published by the MDZ Digital Library team (dbmdz)\footnote{ \url{https://huggingface.co/dbmdz/bert-base-german-cased}.}, which has the same architectural setup as the English BERT$_{\text{BASE}}$ counterpart\footnote{$12$ multi-head attention layers with $12$ attention heads per layer and $768$-dimensional output embeddings.} and is pre-trained on a large corpus of German news reports, books and Wikipedia articles. We initialize all of KPI-Check's trainable parameters randomly from a normal distribution $\mathcal{N}(0,0.02)$ and fix the same random seed of $42$ for all training runs. In addition, we utilize the AdamW \cite{loshchilov2017decoupled} optimizer with a linear warmup of $10$\% and a linearly decaying learning rate schedule. Further, we apply weight decay of $0.01$, clip gradients by normalizing their length to $1$ and set the width embedding dimension of $\mat{w}^\text{width}$ to $25$. In line with Table \ref{tab:grid_search} we also evaluate different levels of dropout regularization, various peak learning rates, batch sizes and hidden dimensions.

\begin{table}[t]
\centering
\input{tables/grid-search}
\caption{Evaluated hyperparameter configurations of KPI-Check's sub-modules, the filtering component and the contrastive autoencoder (CAE) classification head. The best configuration on the validation set is highlighted in boldface. The classification thresholds $\alpha_1$ and $\alpha_2$ are tuned in the $[0,1]$ interval based on the best validation set micro F$_1$-score performance. 
For details on KPI-BERT we refer to \cite{hillebrand2022kpi}.
}
\label{tab:grid_search}
\end{table}

Our experiments are conducted on four Nvidia Tesla V100 GPUs and the model plus training code is implemented in PyTorch. We train the BERT-based filtering model for $10$ epochs and find the best performing model on the validation set after epoch $9$ using early stopping. The total training time amounted to $52$ hours and $14$ minutes. The final KPI matching classification network trained for $15$ epochs until convergence with a training time of $7$ hours and $3$ minutes.

\begin{table*}[t]
     \centering
     \input{tables/kpi-results}
     \caption{Test set results of the sentence/table pair filtering sub-task (a) and the final task of matching semantically KPIs (b). Our full model, KPI-Check, achieves the highest micro- and macro F$_1$ scores of $73.00$\% and $70.52$\%, which significantly improves upon the variation with no filtering module and outperforms the other baselines which all employ the filtering module. 
     We also report upper bound metrics for our approach, assuming a perfect filtering module with no mistakes.}
     \label{tab:kpi-results}
\end{table*}

\subsection{Baseline and Ablations}

We compare the fine-tuned setup of KPI-Check with three baselines and an additional variation discarding the filtering module.
Each baseline makes use of KPI-BERT to first extract and link relevant KPIs and their numerical values from sentences. Also, all competing methods are fine-tuned individually on the validation set and only their respective best setup is evaluated on the hold-out test set.

First, we establish a simple baseline, which we denote Fuzzy String Matching, that drops all learnable components and solely employs fuzzy string matching to link semantically equivalent KPIs. Concretely, it utilizes the extracted KPI strings from sentences (KPI-BERT) and tables and links them by applying the weighted string similarity function \texttt{WRatio()} from the \texttt{rapidfuzz} Python library which is based on the Levenshtein distance \cite{levenshtein1965binary}. If the similarity exceeds a fixed threshold which is tuned on the validation set, we classify two candidate KPIs as semantically equivalent.

Second, we benchmark KPI-Check and its CAE classification layer depicted in Figure \ref{fig:kpi-check}c against an MLP with two fully-connected layers. It is enclosed by relu activation functions and dropout and is followed by a sigmoidal output layer. Formally, the MLP is defined as
\begin{align}
    \hat{y} = \text{MLP}\left(\left[\mat{\bar{s}};\mat{\bar{t}}\right]\right).
\end{align}

Third, we compare KPI-Check's classification performance with a siamese network, first introduced by \cite{bromley1993signature},  that is also trained via contrastive learning but utilizes the siamese architecture of two identical MLPs with shared weights. In particular, the sentence- and table KPI embeddings $\mat{\bar{s}}$ and $\mat{\bar{t}}$ are passed individually through the same MLP and the similarity of their resulting representations is calculated based on the cosine similarity. Similar to the CAE we employ a contrastive loss to minimize the cosine similarity for unrelated KPI-pairs while maximizing it for semantically equivalent pairs. During inference the classification decision is thus based on the cosine similarity score and a fixed threshold which is again tuned on the validation set.

Note both the MLP and the Siamese Network make use of the previous filtering module to ensure a fair comparison to KPI-Check.


Last, we train a fine-tuned ablation version of our complete system that discards the filtering module and directly encodes and classifies all KPI-pairs. We call this variation KPI-Check$_\text{no filtering}$.

\subsection{Results} \label{sec:results}

We evaluate and compare KPI-Check, its variations and all baselines on the previously specified hold out test set.
Table \ref{tab:kpi-results} reports micro- and macro precision, recall and F$_1$ scores for (a) the filtering sub-task and (b) the final task of matching semantically equivalent KPIs. 

First, it can be seen that classifying learned KPI representations significantly outperforms the purely string-based approach of fuzzy string matching by almost more than $25$ percentage points on all metrics.

Second, we find that including a carefully tuned filtering module improves KPI-Check's overall micro and macro F$_1$ score performance by $4$ and $5$ percentage points to $73.00$\% and $70.52$\%, respectively. The same filtering module is also utilized by the baselines, MLP and Siamese Network. However, KPI-Check with its contrastive autoencoder classification head outperforms both competing approaches.

Since the filtering of irrelevant sentence/table pairs is applied before the actual KPI-pair classification, any errors are inevitably carried over to the KPI matching module. Hence, we analyze different filtering thresholds $\alpha_1$ to find the right trade-off between reducing the dataset imbalance on the one hand and falsely removing correct KPI-pairs on the other hand. The best performing model is included in KPI-Check and the other baselines and it is separately evaluated on the filtering sub-task (see Table \ref{tab:kpi-results}a). Further, we report KPI-Check's hypothetical performance, assuming a perfect filtering model that makes no mistakes. These numbers can be interpreted as an upper bound for the current approach.

\begin{table}[t]
     \centering
     \input{tables/filtering-statistics}
     \caption{Impact of sentence/table pair filtering on the KPI-pair imbalance. Our actual filtering model drastically reduces the number of negative KPI-pairs while keeping the majority of positive once to reduce the overall test set imbalance from $395:1$ to $46:1$. ``$-$'' denotes no filtering (see Table \ref{tab:data-statistics})
     and ``perfect'' assumes a perfect filtering model that makes no mistakes to establish an upper bound.}
     \label{tab:filtering-statistics}
 \end{table}
Table \ref{tab:filtering-statistics} shows the impact of our filtering module on the dataset imbalance of positive and negative KPI-pairs. While the number of negative test set pairs decreases considerably from around $2.1$ million to $0.2$ million ($91$\%), the number ob positive pairs remains relatively high with $4,408$ compared to originally $5,431$ ($19$\% decrease). Of course, the difference of $1,023$ wrongly filtered out pairs is in the process automatically classified negatively, thereby reducing the maximum possible micro recall to $81,16$\%. 

\begin{table*}[t]
     \centering
     \input{tables/linking-examples}
     \caption{Translated test set samples of wrongly annotated but correctly predicted key performance indicator (KPI) pairs. Examples 1 to 3 indicate rounding and unit errors in the financial report leading to wrong negative annotations via automated number matching. Examples 4 and 5 reveal accidentally matched but actually unrelated KPIs.}
     \label{tab:linking-examples}
 \end{table*}
In addition to the quantitative evaluation, Table \ref{tab:linking-examples} qualitatively highlights a few test set examples where KPI-Check correctly predicts the equivalence of two KPIs, whereas the automated annotations generated via number matching (see Section \ref{sec:number-matching}) are wrong. Examples 1 to 3 reveal rounding and unit inconsistencies in the current and prior year values which lead to wrong negative annotations. Nevertheless, KPI-Check correctly predicted these samples as semantically equivalent. In the contrary, examples 4 and 5 show cases of accidental current and prior year value matching that result in wrong positive annotations. However, the model again correctly classified both samples as completely unrelated.

In summary, these examples demonstrate the model's generalization capability despite automatically generated imperfect annotations.

\section{Conclusion and Future Work}

Numerical inconsistencies of key performance indicators (KPIs) within published financial reports may diminish investors' trust in a firm's compliance and governance process which potentially impacts the firm economically and in the worst case harms its reputation.

In this work, we approach this issue by introducing KPI-Check, a novel system that aids auditors in automatically identifying semantically equivalent KPIs and validate their numerical consistency. The architecture combines a tailored financial named entity and relation extraction module with a BERT-based filtering component and  a contrastive autoencoder (CAE) based text pair classification head. It first extracts KPIs and their numerical facts from unstructured sentences before linking them to synonymous mentions in the balance sheet and profit \& loss statement. It achieves a strong matching performance of $73.00$\% micro F$_1$-score on a hold-out test set, which shows the model's capability to detect semantically equivalent KPIs with a high confidence.

KPI-Check is momentarily being integrated in the auditing process of a major auditing company and first user tests have already promised significant efficiency gains.

In future work, we plan to extend the system to arbitrary table types to guarantee automated KPI cross-checking across the entire financial document. While the balance sheet and profit \& loss statement arguably represent the most important pieces of information, many less relevant KPIs are reported in smaller tables throughout the document. Semantically linking these to other table- and sentence occurrences is high on our agenda.

Further, we set out together with our industry partner to manually annotate a small number of financial statements with respect to creating high quality ground truth labels of semantically equivalent KPIs. Currently, our model training and quantitative evaluation builds on a carefully designed automated annotation process leveraging informed number matching. Despite the high annotation quality and the model's success we hope to improve the quantitative evaluation even further by utilizing manually crafted labels.

\section{Acknowledgment}

This research has been funded by the Federal Ministry of Education and Research of Germany and the state of North-Rhine Westphalia as part of the Lamarr-Institute for Machine Learning and Artificial Intelligence, LAMARR22B.

\renewcommand*{\bibfont}{\footnotesize}
\printbibliography

\end{document}

%% file: tables/pandl-example.tex
\rowcolors{2}{gray!10}{white}
\begin{tabular}{lrr}
\toprule
\rowcolor{white} in \$ &  2019/2020 & 2018/2019 \\
\midrule
1. Gross profit             & $24,059,512.21$ & $22,051,698.38$ \\
 & & \\
2. Personnel expenses       & &  \\
a) Wages and salaries             & $15,675,943.67$ & $13,231,237.73$ \\
b) Social security contributions   & $1,375,421.49$ & $1,865,432.63$ \\
\dots & \dots & \dots \\
\bottomrule
\end{tabular}

%% file: tables/data-statistics.tex
\begin{tabular}{lrrr}
\toprule
 & Training & Validation & Testing \\
\midrule
Documents            & $6,032$ & $764$ & $752$ \\
\midrule
KPI Pair Statistics  & & & \\
\hspace{1em}Total         & & & \\
\hspace{2em}Pairs $+$            & $48,736$	&		$5,893$ &			$5,431$ \\
\hspace{2em}Pairs $-$            & $17,871,922$	&		$2,192,954$ &			$2,145,354$ \\
\hspace{2em}Imbalance            & $367:1$	&		$372:1$ &			$395:1$ \\
\hspace{1em}Average per Document         & & & \\
\hspace{2em}Pairs $+$            & $8$	&		$8$ &			$7$ \\
\hspace{2em}Pairs $-$            & $2,963$	&		$2,870$ &			$2,853$ \\
\hspace{2em}Imbalance            & $510:1$	&		$492:1$ &			$560:1$ \\
\midrule
\multicolumn{2}{l}{Sentence/Table Pair Statistics}  & & \\
\hspace{1em}Total         & & & \\
\hspace{2em}Pairs $+$            & $42,912$	&		$5,218$ &			$4,812$ \\
\hspace{2em}Pairs $-$            & $503,150$	&		$64,423$ &			$60,543$ \\
\hspace{2em}Imbalance            & $12:1$	&		$12:1$ &			$13:1$ \\
\hspace{1em}Average per Document         & & & \\
\hspace{2em}Pairs $+$            & $7$	&		$7$ &			$6$ \\
\hspace{2em}Pairs $-$            & $83$	&		$84$ &			$81$ \\
\hspace{2em}Imbalance            & $17:1$	&		$16:1$ &			$17:1$ \\
\bottomrule
\end{tabular}

%% file: tables/grid-search.tex
\scriptsize
\begin{tabular}{llc}
\toprule
Sub-Module & Hyperparameter &  Configurations \\
\midrule
\multirow{5}{*}{\makecell{BERT$_\text{filter}$ \\ + MLP$_\text{filter}$}}   & Batch size      & $2$, $\bm{4}$, $8$ \\
                                                              & Learning rate   & $1e^{-4}$, $\bm{1e^{-5}}$, $1e^{-6}$ \\
                                                              & Dropout         & $0.0$, $\bm{0.1}$, $0.2$, $0.3$ \\
                                                              & Confidence threshold $(\alpha_1)$ & $[0,\bm{0.0035},1]$ \\
                                                              & MLP hidden dimensions & \makecell{\textbf{no}, $(1024, 128)$, $(2048, 256)$, \\ $(2048, 1024, 256)$} \\
\midrule
\multirow{6}{*}{CAE}      & Batch size      & $32$, $\bm{64}$, $128$ \\
                                       & Learning rate   & $1e^{-4}$, $\bm{1e^{-5}}$, $1e^{-6}$ \\
                                       & Dropout         & $0.0$, $\bm{0.1}$, $0.2$, $0.3$ \\
                                       & Confidence threshold $(\alpha_2)$ & $[0,\bm{0.7099},1]$ \\
                                       & Hidden dimensions (enc/dec) & \makecell{ $(1024, 128)$, $\bm{(2048, 256)}$, \\ $(2048, 256, 64)$} \\
                                        & margin ($m$) & $0.5$, $\bm{1.0}$ \\

\bottomrule
\end{tabular}

%% file: tables/kpi-results.tex
\begin{tabular}{llcccccc}
\toprule
in \% & & \multicolumn{3}{c}{Micro} & \multicolumn{3}{c}{Macro} \\
\cmidrule(lr){3-5}
\cmidrule(lr){6-8}
Task & Architecture & Precision &  Recall &  F$_1$ &  Precision &  Recall & F$_1$ \\
\midrule
(a) Filtering & $\text{BERT}_\text{filter} + \text{MLP}_\text{filter}$ & $70.72$ &     $80.01$ &     $75.08$ & $72.75$  &   $81.73$ & $70.31$ \\
\midrule
\multirow{6}{*}{(b) KPI Matching} & Fuzzy String Matching    & $49.18$       &  $40.88$  &   $44.65$ & $56.91$  &   $48.62$  &   $41.37$ \\
& Siamese Network       & $52.94$       &  $67.43$ &   $59.31$ & $64.15$  &   $72.19$  &   $60.98$ \\
& MLP       & $71.83$  &  $73.71$  &   $72.76$  & $76.09$   &  $76.76$   &  $70.21$ \\
& KPI-Check$_\text{no filtering}$       & $62.13$       &  $\bm{77.55}$ &   $68.99$ & $66.42$  &   $\bm{80.34}$  &   $65.71$ \\
& KPI-Check                             & $\bm{73.16}$  &  $72.84$  &   $\bm{73.00}$  & $\bm{77.21}$   &  $76.21$   &  $\bm{70.52}$ \\
\cmidrule(lr){2-8}
& KPI-Check$_\text{perfect filtering}$  & $90.81$ &    $84.09$ &    $87.32$ &  $91.79$  &   $83.65$ &    $85.57$\\
\bottomrule
\end{tabular}



%% file: tables/filtering-statistics.tex
\begin{tabular}{lcrrr}
\toprule
KPI Statistics & Filtering & Training & Validation & Testing \\
\midrule
\multirow{3}{*}{Pairs $+$} & $-$           & $48,736$	&		$5,893$ &			$5,431$ \\
                     & model          & $48,731$	&		$4,659$ &			$4,408$ \\
                     & perfect            & $48,736$	&		$5,893$ &			$5,431$ \\
  \midrule
\multirow{3}{*}{Pairs $-$} & $-$          & $17,871,922$	&		$2,192,954$ &			$2,145,354$ \\
                     & model         & $1,688,112$	&		$199,468$ &			$200,839$ \\
                     & perfect            & $1,589,333$	&		$189,152$ &			$180,401$ \\
  \midrule
\multirow{3}{*}{Imbalance} & $-$           & $367:1$	&		$372:1$ &			$395:1$ \\
                     & model         & $35:1$	&		$43:1$ &			$46:1$ \\
                     & perfect            & $32:1$	&		$32:1$ &			$33:1$ \\
\bottomrule
\end{tabular}

%% file: tables/linking-examples.tex

\begin{tabular}{llr@{\hspace{0.25em}}lr@{\hspace{0.25em}}lcc}
\toprule
 & KPI-Pair (Text/Table) & \multicolumn{2}{c}{Current Year} & \multicolumn{2}{c}{Prior Year} & Prediction & Label \\
\midrule
\multirow{2}{*}{1} & Active difference         & 991 & T\euro{}	&	&	 &			\multirow{2}{*}{$+$} & \multirow{2}{*}{$-$} \\
 & E. Active difference from asset offsetting         & 992 & T\euro{}	& 863 & T\euro{}		 &			 &  \\
\midrule
\multirow{2}{*}{2} & Net income         & 293,475.51 & \euro{}	& 1,078,302.11	& T\euro{}	 &			\multirow{2}{*}{$+$} & \multirow{2}{*}{$-$} \\
 & 15. Net income     & 293,475.51 & \euro{}	& 1,078,302.11  & \euro{}		 &			 &  \\
\midrule
\multirow{2}{*}{3} & Passive surplus of deferred taxes         & 170 & T\euro{}	& 226	& T\euro{}	 &			\multirow{2}{*}{$+$} & \multirow{2}{*}{$-$} \\
 & G. Passive deferred taxes     & 170,500 & \euro{}	& 226  & T\euro{}		 &			 &  \\
\midrule
\multirow{2}{*}{4} & Supervisory board remuneration         & 16 & T\euro{}	& 16	& T\euro{}	 &			\multirow{2}{*}{$-$} & \multirow{2}{*}{$+$} \\
 & Other provisions     & 15,710 & \euro{}	& 15,710  & \euro{}		 &			 &  \\
 \midrule
\multirow{2}{*}{5} & Reserve for own shares         & 1,499 & T\euro{}	& 	& 	 &			\multirow{2}{*}{$-$} & \multirow{2}{*}{$+$} \\
 & 21. Withdrawals from other revenue reserves     & 1,499 & T\euro{}	& 390  & T\euro{}		 &			 &  \\
\bottomrule
\addlinespace[1ex]
\multicolumn{8}{l}{\scriptsize{T $=$ thousand}}
\end{tabular}